# Tunnel Facility-based Vehicle Localization in Highway Tunnel using 3D LIDAR

Kyuwon Kim, Junhyuck Im, and Gyuin Jee, *Member, IEEE*

*Abstract*—Vehicle localization in highway tunnels is a challenging issue for autonomous vehicle navigation. Since GPS signals from satellites cannot be received inside a highway tunnel, map-aided localization is essential. However, the environment around the tunnel is composed mostly of an elliptical wall. Thereby, the unique feature points for map matching are few unlike the case outdoors. As a result, it is a very difficult condition to perform vehicle navigation in the tunnel with existing map-aided localization. In this paper, we propose tunnel facility-based precise vehicle localization in highway tunnels using 3D LIDAR. For vehicle localization in a highway tunnel, a point landmark map that stores the center points of tunnel facilities and a probability distribution map that stores the probability distributions of the lane markings are used. Point landmark-based localization is possible regardless of the number of feature points, if only representative points of an object can be extracted. Therefore, it is a suitable localization method for highway tunnels where the feature points are few. The tunnel facility points were extracted using 3D LIDAR. Position estimation is conducted using an EKF-based navigation filter. The proposed localization algorithm is verified through experiments using actual highway driving data. The experimental results verify that the tunnel facility-based vehicle localization yields precise results in real time.

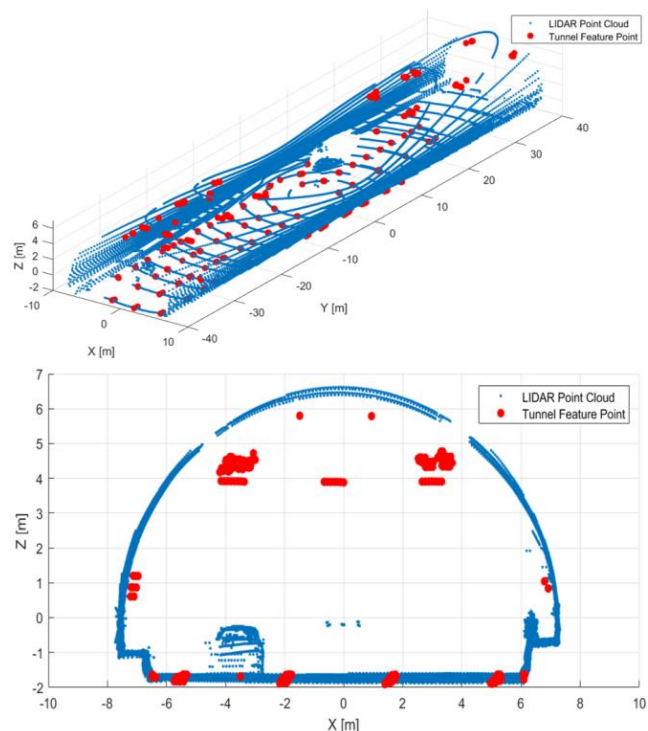

Fig. 1. Environment of highway tunnel scanned by 3D LIDAR. The blue dots are LIDAR point cloud, and red dots are tunnel feature points extracted from tunnel facilities.

## I. INTRODUCTION

VEHICLE localization is one of the most important tasks in autonomous driving. In general, for autonomous vehicles to perform lane-level positioning, the position must be estimated within 0.5 m in the lateral direction [1]. To achieve this, the initial version of the autonomous vehicle estimated the ego-vehicle's position using an expensive GPS (RTK)/inertial navigation system (INS). However, in a GPS-denied environment such as an urban canyon, a large position error occurs owing to the multi-path. Consequently, the lane-level positioning condition cannot be satisfied in this area. To solve this problem, research is being conducted to estimate the position by applying a precision map generated using the mobile mapping system (MMS) [2]. As a result, several studies determined position estimation results to enable lane-level positioning in urban canyons. However, research on vehicle localization in an environment where GPS signals cannot be received for a long time, such as a highway tunnel, is insignificant.

In this paper, we propose a precise vehicle localization algorithm for highway tunnels using 3D LIDAR. Unlike in general outdoor environments, GPS satellite signals cannot be received in tunnels. In addition, most highway tunnels are long because they penetrate mountains. Therefore, the drift error increases with time even when a precise dead reckoning (DR) system is used, thereby rendering lane-level positioning unfeasible. Consequently, map-aided localization is essential.

3D LIDAR is widely used for map-aided localization because it provides precise distance and intensity information. There are mainly two types of map matching using 3D LIDAR. One is the scan matching method based on the iterative closest point (ICP) [3], [4] or normal distribution transform (NDT) [5], [6]. The other is to convert the point cloud into a grid and perform correlation matching [7]. In these methods, the entire LIDAR point cloud or extracted feature points in the point

K. Kim, and G. Jee are with the Navigation and Control System Laboratory, Department of Electrical and Electronics Engineering, Konkuk University, Seoul 05029, Korea (e-mail: gijee@konkuk.ac.kr).
J. Im is with the Spatial Information Research Institute, Korea Land and Geospatial Informatix Corporation, Jeollabuk-do 55365, Korea.

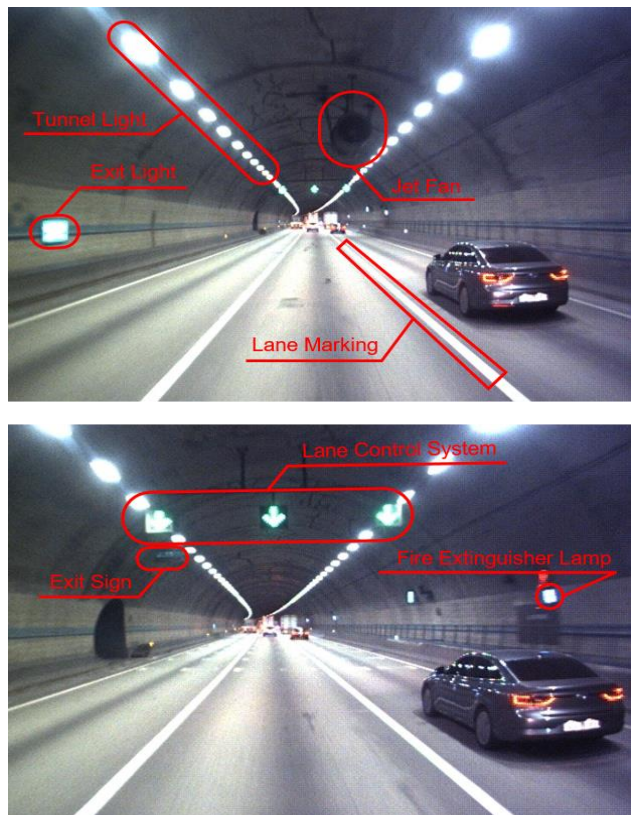

Fig. 2. Types of tunnel facilities. These facilities are infrastructures for safe driving and evacuation. Their installation is mandatory.

cloud are matched with a map. These methods have good matching performance when there are many feature points that are distinguishable from other regions in the scanned scene. In urban canyons or outdoor roads, feature points can be obtained from road-marking [8], [9], vertical structures (such as poles [10], traffic signs [11], and outer walls of buildings [12]). Therefore, precise vehicle localization performance can be obtained similarly as above. However, in a highway tunnel, it is difficult to obtain good performance with the above method because the same scene is repeated.

Fig. 1 shows the environment of a highway tunnel scanned by 3D LIDAR. As shown in Fig. 1, most of the LIDAR points are scan data of an elliptical wall and ground. Meanwhile, the number of points of protruding structures on the wall is marginal. As described above, map matching is ineffective when the number of unique feature points that are distinguishable from other scenes is marginal. This degrades localization performance. In addition, lane marking points extracted from the ground are feature points that are commonly used for vehicle localization. Various road markings are present in outdoor environments, and lane markings are generally dashed lines. Therefore, precise vehicle localization is possible with only road markings. However, in tunnels, it is unfeasible to estimate the longitudinal direction because the lane marking is a solid line. This hinders the precise estimation of vehicle position when only lane marking is used.

To solve this problem, we perform vehicle localization based on a point landmark map [13] and probability distribution map [14]. The point landmark map stores the center points of the tunnel facilities, and the probability distribution map stores the mean and covariance of the lane marking.

Tunnel facilities are installed for safe vehicle driving and evacuation in case of an accident. The types of tunnel facilities are shown in Fig. 2. Point landmark map-based localization is conducted through map matching by detecting the center points of objects in a point cloud. Therefore, regardless of the number of feature points, the localization performance is good if many representative object points are detected. A disadvantage of point landmark map-based localization is that false detections and mismatches degrade the localization performance. However, most tunnel facilities are installed protruding from the wall and at heights set by regulations. Thus, with the removal of only the outer wall of the tunnel, the facilities can be detected conveniently by setting the region of interest (ROI) for the height. In addition, tunnel facilities are installed at large intervals (at least 50 m), and the installation height is different for different types of facilities (Fig. 2). Therefore, mismatch does not occur if the facilities are stored in a map according to their type. Consequently, a point landmark map is suitable for vehicle localization in highway tunnels.

Probability distribution maps are frequently used when matching with road-marking points. Map matching using a probability distribution map (which is point-to-probability distribution matching) is faster than ICP scan matching (which is point-to-point matching). Moreover, as the shape of the object and the probability distribution are similar, the matching performance is higher. Hence, it is a highly suitable map to express lane marking, which is a simple solid line.

The remainder of this paper is organized as follows: Section II introduces related works on vehicle localization, including those involving tunnels. Section III presents a system overview of the experimental vehicle platform and point landmark map. Section IV explains the method for extracting tunnel facility points and analyzes each type of tunnel facility for vehicle localization. Section V describes a tunnel facility-based vehicle localization algorithm. In Section VI, the proposed algorithm is verified experimentally. Finally, Section VII concludes this paper.

## II. RELATED WORKS

Map-aided localization is performed by matching a map and sensor data. The sensors that are mainly used for map-aided localization are LIDAR and cameras. The precise vehicle position is determined by correcting the error of the GPS/DR through map matching. Many studies have stored information on facilities that are generally observed in the vehicle driving environment in a map and have utilized it for vehicle localization.

Hata and Wolf [15] stored curbs and road markings in a map. These were extracted using 3D LIDAR. The vehicle position was estimated using a particle filter-based localization method. The road markings were extracted conveniently by the intensity of LIDAR. Furthermore, the range of particles was limited effectively by curbs to obtain precise position estimation.

In [16], the authors proposed a vehicle localization method based on a map called an extended line map. It stores road markings and building outer walls as lines. The stored line was converted into an occupancy grid map, and map matching was performed through a FFT-based correlation. Thereby, precise vehicle localization results were obtained at high processing speeds in urban canyons.

Choi *et al.* [17] presented a vehicle localization method that uses a camera and is based on a digital map that stores the end points of lane marking and vertices of road signs. Vehicle localization is performed based on a particle filter. The weight of the particles sampled for each lane is calculated through the end points of the lane marking. Furthermore, when road sign vertices are detected, the final lane is determined to estimate the vehicle position on the highway. In this research, precise position estimation was achieved in the highway scenario using a camera. However, it can be observed that because the lane marking is a solid line in the tunnel section, the longitudinal error increases.

As mentioned above, many studies have precisely estimated vehicle position in outdoor environments through map-aided vehicle localization. However, few studies have estimated vehicle position in highway tunnels. As shown in Daoust et al. [18] and Duff et al. [19], studies have performed position estimation in tunnels using LIDAR. However, these methods were appropriate for position estimation in subways and mine tunnels, respectively, and therefore, unsuitable for vehicle localization in highway tunnels.

In general, the vehicle position in highway tunnels is estimated by installing additional infrastructure, such as ultra-wideband (UWB) that is used for indoor positioning [20], [21]. In [22], vehicle localization in highway tunnels through vehicle-to-vehicle (V2V) and vehicle-to-infrastructure (V2I) communication is proposed. Particle filter-based vehicle localization is performed through range measurement transmitted by an impulse radio ultra-wideband (IR-UWB) installed in vehicles and tunnels. However, this requires the installation of additional infrastructure (such as IR-UWB) in the tunnel, for vehicle localization.

In this study, the point landmark map is generated using the tunnel facilities present in a tunnel and without installing additional infrastructure. In our previous work [23], vehicle localization was performed using only one type of tunnel facility: fire extinguisher lamp. Fire extinguisher lamps are installed at intervals of 50 m, which corresponds to the highest density among the tunnel facilities. In addition, these can be detected conveniently through the ROI setting if the tunnel wall is removed effectively. Therefore, it is suitable for use in vehicle localization. However, false detection is likely because the size of fire extinguisher lamp is not large. Moreover, because only one facility is used, the availability is reduced substantially if the fire extinguisher lamps are not detected. Therefore, we increase the availability of the point landmark by adding another tunnel facility and thereby, obtain more precise vehicle position estimation results.

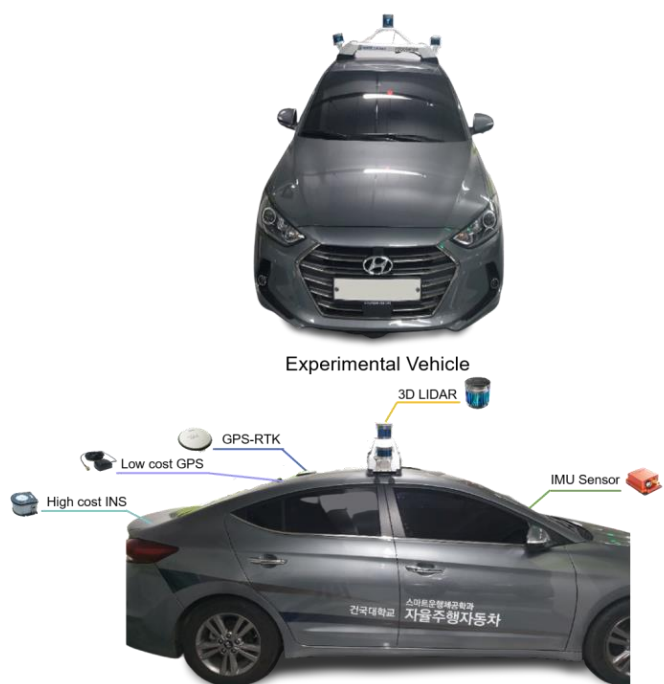

Fig. 3. Configuration of experimental vehicle. The 3D LIDAR (Robosense RS-LIDAR-32), low cost GPS (ublox NEO-M8P), and IMU (MTi-G-700) are used for vehicle localization. GPS-RTK and an expensive INS are used as the reference trajectory.

III. SYSTEM OVERVIEW

*A. Experimental Vehicle Configuration*

The configuration of the experimental vehicle is shown in Fig. 3. The 3D LIDAR (Robosense RS-LIDAR-32) is mounted on top of the vehicle to detect the tunnel facility. A low cost GPS (u-blox NEO-M8P) and IMU (Xsens MTi-G-700) are used to acquire absolute position and DR. In addition, LIDAR calibration [24] is performed using the increments determined through the DR system. LIDAR provides a point cloud by scanning the surroundings through 360° rotation. If the LIDAR moves during the scan interval, the scan data are distorted by the distance moved. In particular, because the vehicle is driven at a high speed on a highway, this distortion is more pronounced. Therefore, we perform LIDAR calibration to compensate for the distortion of the scan data by using DR increments determined by the IMU. GPS-RTK and an expensive INS (Novatel SPAN HG1700) are mounted to determine the ground truth. The sensor specifications for vehicle localization are shown in Table I.

*B. Map Configuration*

In this study, a point landmark map and lane marking map are used for map-aided localization. A point landmark map is a map that stores an extracted representative point of each object. For vehicle localization in a tunnel, we store the center point of the tunnel facility, in the map. The position of the tunnel facility is determined through precision surveying. Unlike other tunnel facilities, lane markings cannot be extracted as landmark points because these are solid lines. Thus, we extracted lane marking vectors from a high-definition map provided by the National Geographic Information Institute in South Korea. Then, we converted the lane marking vectors into probability distributions with a certain length [14]. These are stored as a map. The generated point landmark and lane marking maps are shown in Fig. 4. An example of the content stored in the map is shown in Table II.

TABLE I
SPECIFICATIONS OF SENSORS FOR VEHICLE LOCALIZATION

| | | | | | |
|---|---|---|---|---|---|
| 3D LIDAR (RS-LIDAR-32) | Vertical channel | 32 | Low cost GPS (ublox NEO-M8P) | Position accuracy | 2.5 m circular error probability (CEP) |
| | Vertical FOV | -25°–15° | | Velocity accuracy | 0.05 m/s CEP |
| | | | | Heading accuracy | 0.3° |
| | Range | 200 m | | Data rate | 10 Hz |
| | Angular resolution (horizontal) | 0.1° | IMU (MTi-G-700) | Noise | Accelerometer 60 μg/$\sqrt{hz}$ |
| | | | | | Gyro 0.01 deg/s/$\sqrt{hz}$ |
| | Angular resolution (vertical) | 0.33° | | Bias | Accelerometer 15 μg |
| | | | | | Gyro 10°/h |
| | Accuracy | ± 3 cm | | Data rate | 40 Hz |
| | Data rate | 10 Hz | | | |

TABLE II
EXAMPLE OF MAP CONTENTS

| | Index | Latitude | Longitudinal | Facility Type | | Index | Latitude | Longitudinal | $\sigma_{x^2}$ | $\sigma_{y^2}$ | $\sigma_{xy}$ |
|---|---|---|---|---|---|---|---|---|---|---|---|
| Point landmark map | 1 | 37.2701688 | 127.1832586 | LCS | Lane marking map | 1 | 37.2730432 | 127.1731381 | 2.36 | -2.75 | 3.55 |
| | ⋮ | ⋮ | ⋮ | ⋮ | | ⋮ | ⋮ | ⋮ | ⋮ | ⋮ | ⋮ |
| | N | 37.2734618 | 127.1789613 | Lamp | | N | 37.2794591 | 127.1812632 | 2.14 | -2.89 | 3.21 |

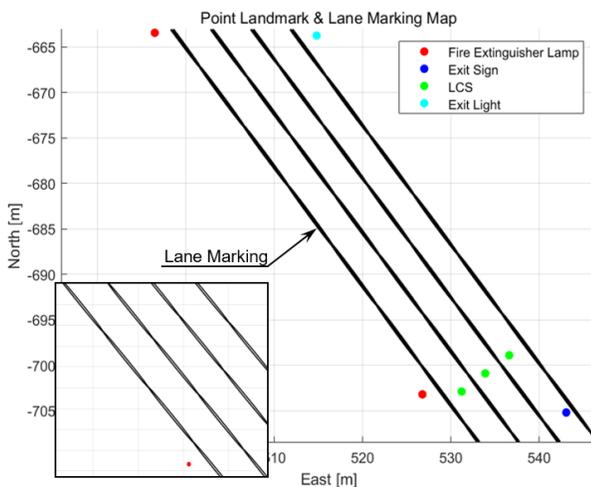

Fig. 4. Point landmark map and lane marking map. The point landmark map stores the center-points of the facilities and is classified by tunnel facility. The mean and variance of the lane marking probability are stored in the lane marking map.

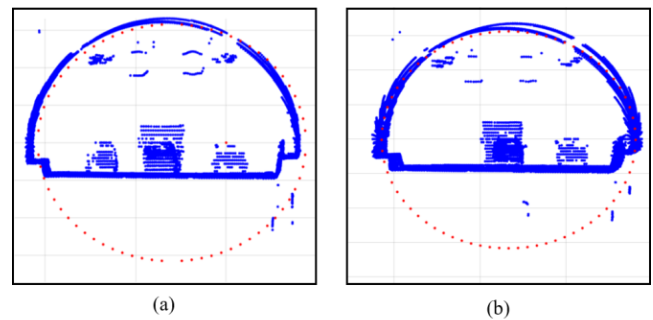

Fig. 5. Result of LIDAR scanned data viewed from vehicle front direction. (a) lateral direction mismatch between vehicle and tunnel frame. (b) azimuth mismatch between vehicle and tunnel frame.

IV. EXTRACTION AND ANALYSIS OF TUNNEL FACILITY

In this section, we explain the method for extracting tunnel facility points using LIDAR. The extracted tunnel facility points are analyzed to select tunnel facilities that can be used for vehicle localization.

*A. Extraction of Tunnel Facility Point*

To match the tunnel facility-based point landmark map, the center point of the tunnel facility must be extracted from the LIDAR data. Lane markings can be extracted conveniently from the point cloud on the road by using the intensity of LIDAR. Meanwhile, other tunnel facilities are attached to the tunnel wall or suspended from the tunnel ceiling. Therefore, the tunnel wall must be removed from the LIDAR scan data to effectively detect these facilities.

Fig. 5 shows the LIDAR scan data viewed from vehicle front direction. As shown in Fig. 5, lateral direction and azimuth mismatches occurred between the vehicle frame and tunnel frame. These inconsistencies occur when the vehicle is horizontally displaced from the center of the lane or when the tunnel wall is curved. If inconsistencies occur, the tunnel wall cannot be removed effectively from the point cloud, and it may cause false detection of tunnel facility points.

To solve this problem, we transform the vehicle frame into the tunnel frame. To achieve this, we create a virtual elliptical cylinder point cloud conforming to the tunnel standard (as shown in Fig. 6). Then, ICP scan matching is performed between the LIDAR scan data and the virtual elliptical cylinder point cloud. The equation for generating a virtual elliptical cylinder point cloud is as follows:

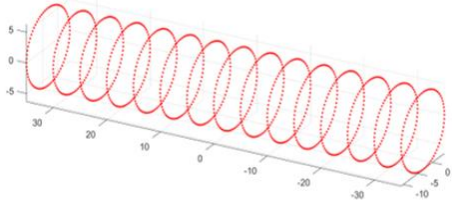

Fig. 6. Point cloud of virtual elliptical cylinder. Each ellipse is created according to the width and height specifications of the tunnel.

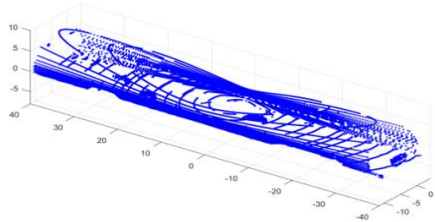

Original Point Cloud

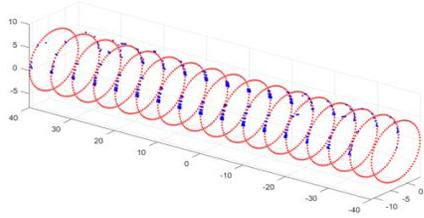

Sampling Result

Fig. 7. Sampling result of point cloud.

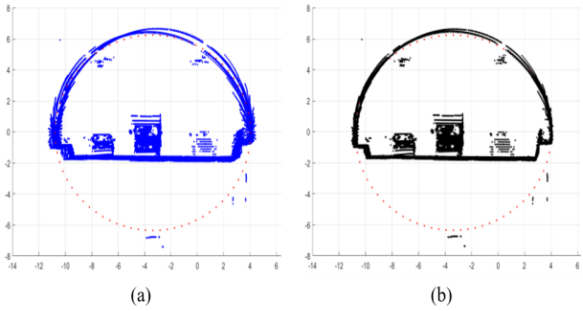

Fig. 8. Conversion to tunnel frame. (a) original point cloud. (b) ICP scan matching result. After ICP scan matching, the azimuth is aligned with the tunnel frame.

$$\frac{x^2}{a^2} + \frac{z^2}{b^2} = 1$$
$$x = \frac{\pm ab}{\sqrt{b^2 + a^2 (\tan\theta)^2}}, \quad z = \pm b\sqrt{1 - (x/a)^2} \quad (1)$$

Equation (1) includes the equation of the ellipse and the coordinates of the point for generating the virtual elliptical cylinder point cloud. $a$ and $b$ are the width and height, respectively, of the tunnel, and $\theta$ is the angular resolution of the elliptical point cloud. In this study, we set the angular resolution as 5°. The points of the ellipse are created using (1). Then, we can create an elliptical cylinder point cloud if we place the points of the ellipse at regular intervals based on the vehicle direction.

For ICP scan matching with the generated elliptical cylinder point cloud, only the points near the tunnel wall are sampled, as shown in Fig. 7. Thus, most of the LIDAR scan points are removed. This increases the processing speed. The matching performance is improved because only the points necessary for ICP matching are sampled.

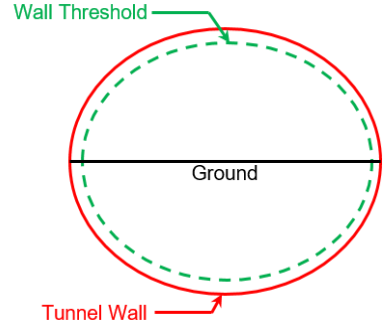

Fig. 9. Example of wall threshold for tunnel wall point cloud removal.

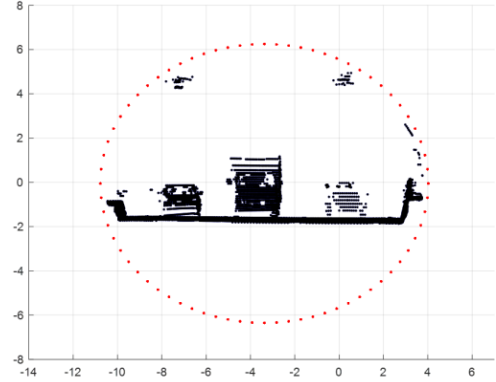

Fig. 10. Removal of tunnel wall point cloud.

Fig. 8 shows the result of the conversion to the tunnel frame. As shown in Fig. 8 (a), the thickness of the tunnel wall is large because of the large angular distortion of the azimuth. Meanwhile, Fig. 8 (b) shows that the azimuth is aligned through ICP scan matching. This reduces the thickness of the tunnel wall. When the point cloud is aligned through ICP scan matching, the tunnel wall must be removed. To remove the tunnel wall point cloud, an ellipse with a size smaller than that of the virtual tunnel point cloud is modeled as shown in Fig. 9, and set as the wall threshold. The wall threshold is used to remove the external points of the elliptical model. The equation for removing the tunnel wall point cloud is as follows:

$$\tilde{a} = a - w_m, \quad \tilde{b} = b - w_m, \quad \theta = \tan^{-1}(z/x)$$
$$\tilde{x} = \frac{\tilde{a}\tilde{b}}{\sqrt{\tilde{b}^2 + \tilde{a}^2 (\tan\theta)^2}}, \quad \tilde{z} = \tilde{b}\sqrt{1 - (\tilde{x}/\tilde{a})^2} \quad (2)$$
$$w_{th} = \frac{\tilde{x}^2}{a^2} + \frac{\tilde{z}^2}{b^2}, \quad \frac{x^2}{a^2} + \frac{z^2}{b^2} < w_{th}$$

Equation (2) expresses the removal of the tunnel wall point cloud. $w_m$ is the wall margin for setting the wall threshold. The width and height of the ellipse are reduced by the wall margin to obtain $\tilde{a}$ and $\tilde{b}$, respectively. Then, the coordinates of the wall threshold ($\tilde{x}$ and $\tilde{z}$) are obtained. The wall threshold $w_{th}$ is determined based on the obtained coordinates. When the ellipse equation is constructed as a scan point, it is considered as a point inside the tunnel if it is smaller than $w_{th}$. If it is larger, the point is removed. It can be observed that, consequently, most of the tunnel wall points are removed (Fig. 10).

TABLE III
LOCATION OF TUNNEL FACILITY

| Tunnel Facility | Location |
| --- | --- |
| Fire extinguisher lamp | Tunnel right wall |
| Exit light | Tunnel left wall |
| Exit sign | Tunnel ceiling adjacent to evacuation corridor |
| LCS | Tunnel ceiling above center of lane |
| Jet fan | Tunnel ceiling |
| Tunnel light | Tunnel ceiling |

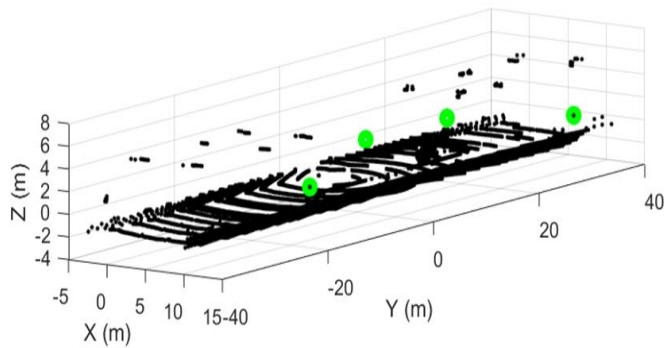

Fig. 11. Result of tunnel facility point extraction. The green dots are extracted tunnel facility points. The tunnel facility points are determined by the location and size of the facility.

After the tunnel wall is removed, the candidate tunnel facilities are extracted through the clustering process. Before clustering, the point cloud is classified by setting the ROI for each tunnel facility. The tunnel facilities have fixed installation locations as shown in Table III [25]. Therefore, the ROI can be set through the installation location for each tunnel facility. This enables the classification of the candidate point cloud for each tunnel facility. Then, clustering is performed using the k-mean clustering algorithm. In clustering, point clouds are clustered for each object. The object is compared with each tunnel facility standard. If the object size satisfies the tunnel facility standard, it is considered as the tunnel facility. The center point of the tunnel facility is extracted by averaging the object points. Fig. 11 shows the result of tunnel facility point extraction. The tunnel facility points are extracted as green dots from the entire tunnel point cloud.

The lane marking is painted on the road unlike the other tunnel facilities. In addition, it is brightly colored for convenient identification by the driver. Therefore, we can extract lane points by the high intensity of the road points. First, the point cloud of the road is extracted with a height filter. Then, the threshold of intensity is set, and the points with an intensity higher than the threshold are extracted as lane marking points.

*B. Analysis of Tunnel Facility for Vehicle Localization*

The candidate tunnel facilities for vehicle localization include lane marking, lane control system (LCS), exit sign, exit light, fire extinguisher lamp, tunnel light, and jet fan. It is essential to install these facilities in highway tunnels. We analyze these facilities in terms of vehicle localization and finally select the tunnel facilities to be used.

Lane marking is painted with a bright color to facilitate its identification. Therefore, it can be conveniently extracted from the LIDAR intensity. However, a lane marking is a solid line in the tunnel, and it is unfeasible to estimate the longitudinal direction. Nevertheless, because lane marking can be used to correct the lateral position error, it can be used as a facility for vehicle localization.

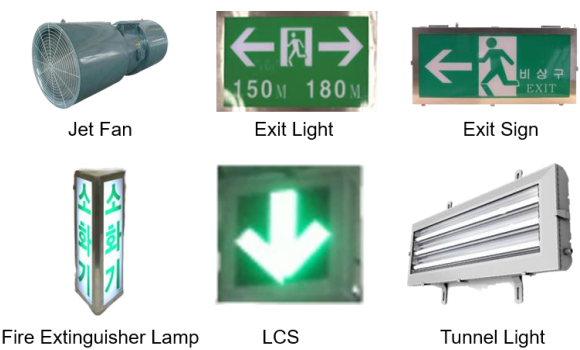

Fig. 12. Candidate tunnel facilities for vehicle localization.

The other tunnel facilities are attached to the tunnel wall or suspended from the tunnel ceiling. Table IV lists the specifications of the tunnel facilities. The complete installation specifications of tunnel facilities can be found in the installation manual for highway traffic safety facilities [26] issued by the Korean Expressway Corporation. The shapes of the tunnel facilities are shown in Fig. 12.

The fire extinguisher lamps are placed at intervals smaller than those for the other facilities. In addition, each lamp protrudes from the tunnel wall to which it is attached. Therefore, it can be detected conveniently by removing the tunnel wall. Similarly, the exit lights are at short intervals (50 m) and can be detected conveniently by removing the tunnel wall. Moreover, the two types of facilities are installed on the left and right walls, respectively. Therefore, even while driving in a lane far from one facility, positioning is possible using the other facility. However, exit lights are installed at a height lower than that of the fire extinguisher lamps and may be obscured by other vehicles. Therefore, the detection rate of the exit lights may be lower than that of the fire extinguisher lamps.

The LCS and exit sign are installed at large intervals. However, these are suspended from the tunnel ceiling and therefore, are at a reasonable distance from the surrounding objects. Consequently, the false detection rate is relatively marginal. In addition, because these are installed facing the vehicle, they play a significant role in the correction of longitudinal error. Moreover, because the installation height is large, the risk of non-extraction owing to being obscured by other vehicles is absent.

Similar to the LCS and exit sign, a jet fan is attached to the tunnel ceiling and installed facing the vehicle. However, the jet fan has a large length, and the LIDAR cannot scan the entire shape of the jet fan. Therefore, a large error in the longitudinal direction is likely if it is stored as a representative point.

The tunnel lights are at intervals of 5–10 m, which corresponds to a density higher than those of the other facilities. The excessive density is likely to result in these lights been matched with other tunnel lights in the map. Moreover, because these are close to other facilities installed in the tunnel ceiling, they could be falsely detected as one of the other facilities.

TABLE IV
SPECIFICATIONS OF TUNNEL FACILITIES

| Tunnel Facility | Interval | Height of Installation | Size (W × L × H) | Remark |
|---|---|---|---|---|
| Fire extinguisher lamp | 50 m | 2.5–3 m | 200 mm × 220 mm × 420 mm | Triangular column shape. |
| Exit light | 50 m | 1.5–2 m | 1200 mm × 30 mm × 730 mm | |
| Exit sign | 200–300 m | 5–5.5 m | 1310 mm × 130 mm × 610 mm | Installed facing the vehicle. |
| LCS | 400–500 m | 5–5.5 m | 800 mm × 250 mm × 800 mm | Installed at 500 m intervals from the tunnel entrance. Installed facing the vehicle. |
| Jet fan | 150–200 m | 5–6 m | 1200 mm × 4900 mm × 1200 mm | Cylindrical shape. Installed facing the vehicle. |
| Tunnel light | 5–10 m | 6–6.5 m | 1400 mm × 400 mm × 150 mm | |

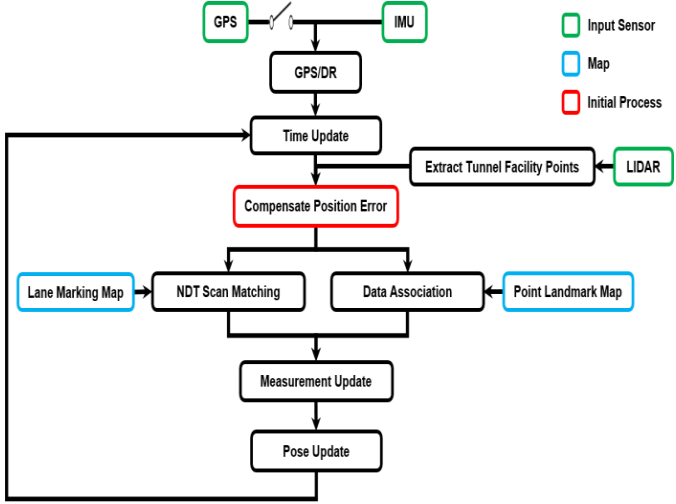

Fig. 13. Flow chart of tunnel facility-based vehicle localization.

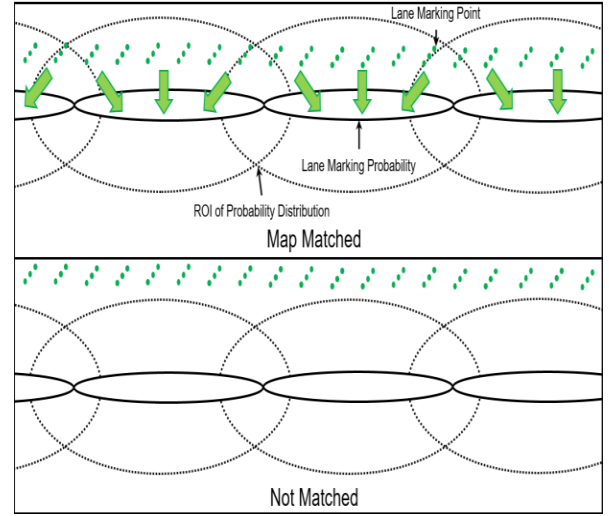

Fig. 14. Example of data association between lane marking point and probability distribution in the lane marking map. Data association is performed by using two distance parameters. One is the Euclidean distance and the other is the Mahalanobis distance.

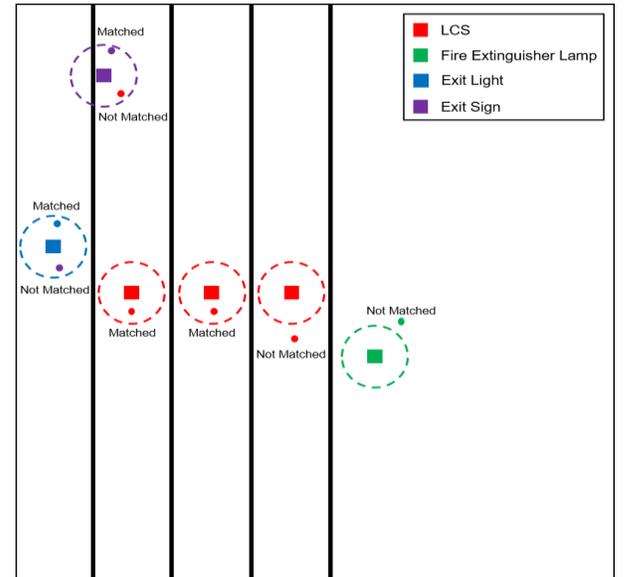

Fig. 15. Example of data association between extracted tunnel facility points and point landmark map points. The extracted tunnel facility points and point landmarks are classified by tunnel facility. If the point enters the range indicated by the dashed line, the map and point are matched.

Consequently, we omit the jet fans and tunnel lights for vehicle localization. The following facilities are selected: lane marking, fire extinguisher lamp, LCS, exit light, and exit sign.

## V. TUNNEL FACILITY-BASED VEHICLE LOCALIZATION

The flow chart of the tunnel facility-based vehicle localization is shown in Fig. 13. The navigation filter for vehicle localization is based on the EKF. GPS satellite signals cannot be received inside the tunnel. Therefore, GPS/DR is performed using the GPS and IMU outside the tunnel, whereas DR is performed using only the IMU in the tunnel. While entering the tunnel, the position error is compensated based on the tunnel facility points extracted from LIDAR. The time update is performed through the speed and yaw rate determined from the DR system. Subsequently, map matching is performed. Lane marking points are matched through NDT scan matching, and the other tunnel facility points are matched through data association. Finally, the vehicle position is determined by performing a measurement update based on the measured values determined by map matching.

### A. Map Matching

There are two maps for performing map matching with tunnel facilities: the lane marking map and point landmark map. The lane marking map consists of probability distributions, and the map matching is performed through NDT scan matching. The other tunnel facility points are matched with a point landmark map through point-to-point data association.

*1) Lane Marking Point Matching*

For NDT scan matching, data association [14] must be performed between a point and a probability distribution. An example of data association between a lane marking point and the probability distribution in a lane marking map is shown in Fig. 14. First, we determine the probability distribution of the closest distance by calculating the Euclidean distance between the lane marking point and the probability distribution of the map. Then, we determine the probability distribution that matches with the lane marking point using the Mahalanobis distance. At this time, we set the ROI of the probability distribution as shown in Fig. 14. We consider data association to be complete when the Mahalanobis distance between the ROI of the probability distribution and the lane marking point is less than a certain threshold.

When data association is complete, the measured value is obtained through NDT scan matching. The measurement of scan matching is a correction value of the position and heading error. NDT scan matching is performed in two steps: 1) Calculate the probability distribution function (PDF) and its Jacobian. The PDF is calculated using the lane marking point and probability distributions of the map. 2) Calculate the score function based on the PDF and use it to determine the error correction value. A detailed description of this method is available in [27].

*2) Tunnel Facility Point Matching*

As explained in Section IV, tunnel facilities can be classified because their installation location and specifications are known. Therefore, it is feasible to separately match each facility while performing data association. An example of data association between extracted tunnel facility points and point landmark map points is shown in Fig. 15. When the tunnel facility point extracted by LIDAR is within the range around the point of the map, we match the extracted tunnel facility and map point. As shown in Fig. 15, we express in the identical color if the tunnel facility point extracted from LIDAR and the facility point of the map are the identical. It can be verified that map matching is performed only when the types of the extracted tunnel facility point and the point on the map are identical. The matched tunnel facility points are used as measurements of the EKF-based navigation filter.

### B. Compensation of Position Error before Map Matching

Data association is an important process for point landmark-based vehicle localization. In addition, data association can be performed only when the extracted point and map point are within a certain range. However, map matching through data association is not performed if the position error is larger than the range. In addition, the drift error of the DR system increases gradually in the tunnel. Consequently, if the position error is not compensated for before map matching, vehicle localization through map matching is unfeasible because the drift error is larger. In particular, because our GPS module is inexpensive, a larger position error may occur. A GPS receiver capable of obtaining RTK correction information can be installed in autonomous vehicles to obtain a more precise position. However, even if we use the RTK system, there may be a variation in position accuracy depending on the environment around the tunnel entrance. Therefore, for stable map matching, it is necessary to compensate for the position error before starting the algorithm.

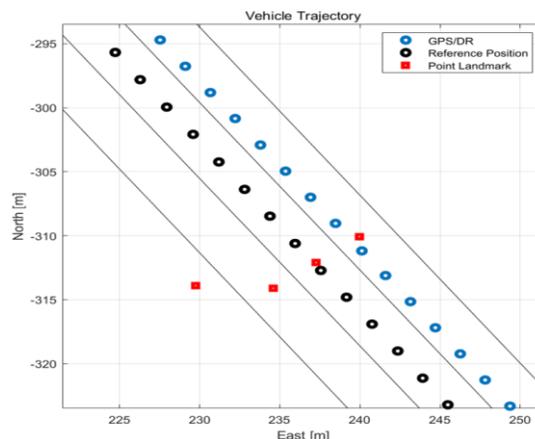

Fig. 16. Vehicle trajectory of GPS/DR and reference position around the tunnel entrance.

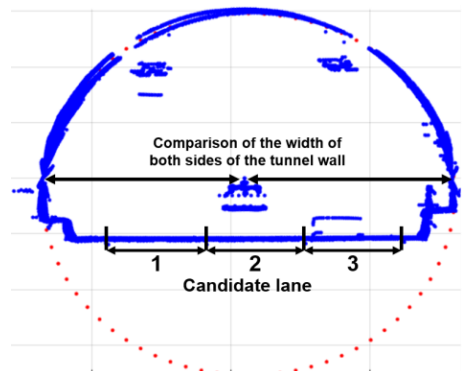

Fig. 17. Example of determination of current lane. The current lane is determined by comparing the widths of both sides of the tunnel wall scanned by LIDAR.

Fig. 16 shows the vehicle trajectory of GPS/DR and the reference position around the tunnel entrance. As shown in Fig. 16, compared with the reference position, the position of the GPS/DR has a large error in the lateral direction. In this case, if the position error is larger than the range of the data association, map matching is unfeasible. To solve this problem, we compensate for the position error before starting the vehicle localization.

*1) Lateral Position*

To correct the lateral error, the lane in which the ego-vehicle is running should be identified. Fig. 17 shows an example of the determination of the current lane. The widths of the tunnel and lanes are constant from the beginning to the end of the tunnel. Therefore, assuming that the number of lanes in the tunnel is known, the current lane can be determined from the distance to the tunnel wall on either side, which is obtained from the LIDAR point cloud, as shown in Fig. 17.

Once the current lane is identified, the position error in the lateral direction is compensated for by using a fire extinguisher lamp point. Fig. 18 shows the result of lateral position error correction using the fire extinguisher lamp point. The fire extinguisher lamp is installed on the right wall of the tunnel, as described in Table III. In addition, it is closest to the tunnel entrance. Therefore, it is the most suitable facility for lateral position error correction.

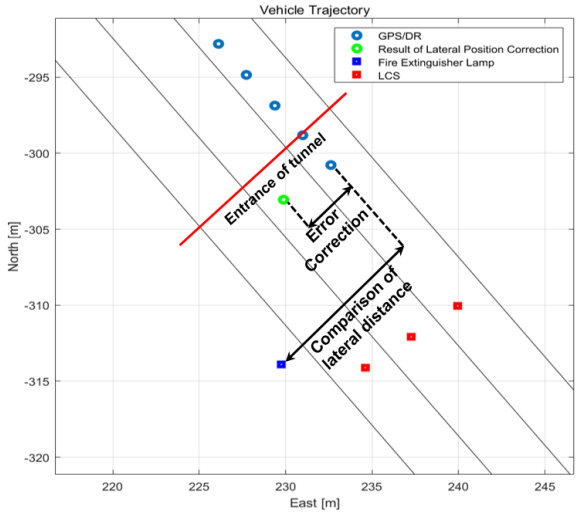

Fig. 18. Result of lateral position error compensation. Error is corrected by a comparison with the lateral distance between the GPS/DR position and fire extinguisher lamp position.

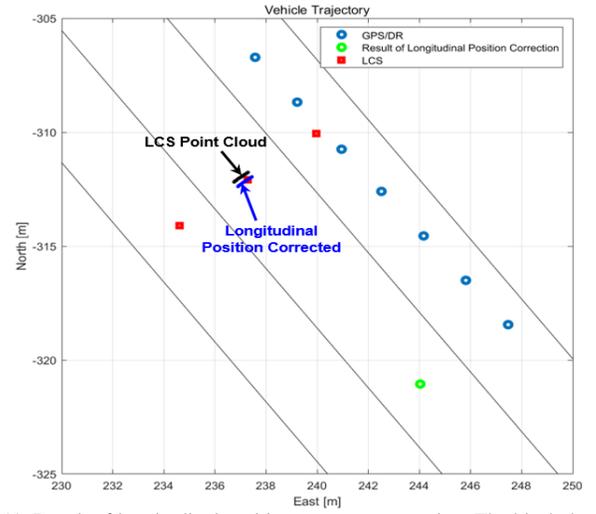

Fig. 19. Result of longitudinal position error compensation. The black dots are LCS points before the longitudinal position error is corrected, and the blue dots are LCS points after the correction.

The lateral position error is corrected through a comparison with the lateral distance between the GPS/DR position and fire extinguisher lamp point. In the tunnel, lane change is prohibited, and the vehicle generally drives in the middle of the lane. Moreover, we know the current lane on which the ego-vehicle is running. Therefore, the lateral position error can be obtained by subtracting the lateral distance between the current lane and fire extinguisher lamp point from that between the GPS/DR and fire extinguisher lamp point.

$$\vec{v}_{DR} = p_{map} - p_{DR}$$
$$\vec{v}_{CL} = p_{map} - p_{CL} \quad (3)$$

Equation (3) is a vector derived by subtracting the position of the fire extinguisher lamp point from those of each point. $\vec{v}_{DR}$ is the vector between the GPS/DR and the fire extinguisher lamp point on the map, and $\vec{v}_{CL}$ is the vector between the center of the current lane and the fire extinguisher lamp point on the map.

$$\vec{l}_{DR} = R^{-1} \cdot \vec{v}_{DR} = [lateral_{DR} \quad longitudinal_{DR}]^T$$
$$\vec{l}_{CL} = R^{-1} \cdot \vec{v}_{CL} = [lateral_{CL} \quad longitudinal_{CL}]^T \quad (4)$$

The lateral distance is derived as $\vec{l}_{DR}$ and $\vec{l}_{CL}$ in the vehicle frame by an inverse rotation matrix $R^{-1}$.

$$\Delta lateral = lateral_{DR} - lateral_{CL}$$
$$\Delta p_{lat} = R \cdot \begin{bmatrix} \Delta lateral \\ 0 \end{bmatrix} \quad (5)$$

From $\vec{l}_{DR}$ and $\vec{l}_{CL}$, a correction value of the lateral position error is determined by the difference between $lateral_{DR}$ and $lateral_{CL}$, which are lateral components. Then, we convert it into a global frame $\Delta p_{lat}$ and add it to the current position. The lateral position is compensated as shown by the green dot in Fig. 18.

*2) Longitudinal Position*

The longitudinal position error correction is performed after the lateral position error correction is completed. The longitudinal position error is compensated using the LCS. It is convenient to estimate the longitudinal direction because the LCS is installed above the center of the lane facing the vehicle. Because the distance from the tunnel ceiling is large, it can be extracted conveniently, and there is a negligible possibility of false detection. In addition, according to the specifications, the LCS is to be installed near the tunnel entrance. This makes it suitable for compensating the longitudinal position error.

Longitudinal position error compensation using LCS is obtained by the difference in the longitudinal distance between the LCS point extracted from the LIDAR point cloud and the LCS point in the point landmark map. However, because each lane has one LCS, we determine the difference in distance using the average of the detected LCS points.

$$\vec{v}_{LIDAR} = p_{map} - p_{LIDAR} \quad (6)$$

Equation (6) is a vector derived by subtracting the position of the LCS point in the map from that of the LCS point extracted from the LIDAR point cloud.

$$\vec{l}_{LIDAR} = R^{-1} \cdot \vec{v}_{LIDAR} = [lateral_{LIDAR} \quad longitudinal_{LIDAR}]^T$$
$$\Delta p_{lon} = R \cdot \begin{bmatrix} 0 \\ longitudinal_{LIDAR} \end{bmatrix} \quad (7)$$

In (7), $\vec{v}_{LIDAR}$ is converted to the vehicle frame $\vec{l}_{LIDAR}$. Subsequently, the longitudinal position error is compensated for by converting only the longitudinal component back to the global frame $\Delta p_{lon}$ and adding it to the current position.

The result of the longitudinal position error compensation is shown in Fig. 19. The black dots are LCS points before the longitudinal position error is compensated, and the blue dots are LCS points after the compensation. As shown in Fig. 19, the corrected longitudinal error is not large. However, the points of the LCS have moved further toward the center of the map point. As a result, the longitudinal error was corrected effectively.

Table V shows the result of the position error compensation.

TABLE V
RESULT OF POSITION ERROR COMPENSATION

|  | Lateral | Longitudinal |
|---|---|---|
| Lane 1 | 0.26 m | 0.21 m |
| Lane 2 | 3.57 m | 0.25 m |
| Lane 3 | 2.02 m | -0.12 m |

As shown in Table V, the position error is over 3 m before map matching. Similarly, when the position error is large, the distance between the map point and the facility point extracted from the LIDAR point cloud increases during data association. Therefore, map matching is not feasible. Thus, a process to compensate for the position error is necessary. After the position error is compensated, vehicle localization is performed using the EKF-based navigation filter.

*C. EKF Configuration*

EKF-based vehicle localization [28] consists of two steps: time update and measurement update. First, we compose a DR system based on the accelerometer and gyro of the IMU. Next, we perform a time update using the speed and yaw rate derived using the DR system. The measurement update is performed using two measurements: 1) the error correction value determined by NDT scan matching between the lane marking points and probability distributions of the map and 2) the matching result obtained by data association between the extracted tunnel facility points from LIDAR and the point landmark map.

*1) Time Update*

The input data for vehicle localization are acquired through the GPS/DR system. The GPS/DR system [29] is composed of a low cost GPS and IMU. Outside the tunnel, the vehicle position and heading are estimated using GPS and IMU measurements. We can acquire the absolute position using GPS, and the increments of position and heading are acquired from accumulated IMU measurements. When the vehicle enters the tunnel, the DR is performed using only the IMU.

For the EKF, it is necessary to define the state and input vector.

$$X_t = [x_t \ y_t \ \psi_t]^T$$
$$u_t = [v_{DR_t} cos(\psi_t) \ v_{DR_t} sin(\psi_t) \ \dot{\psi}_{DR_t}]^T \quad (8)$$

In (8), $X_t$ and $u_t$ are the state and input vectors, respectively. The state vector comprises the vehicle's position and heading. $v_{DR_t}$ and $\dot{\psi}_{DR_t}$ are the vehicle speed and yaw rate estimated from the DR system, respectively. We derive the increment in position through the DR speed and vehicle heading state. The time update is performed as in (9) using the state and input vector.

$$\hat{X}^-_{t+1} = F\hat{X}_t + Gu_t$$
$$P^-_{t+1} = FP_t F + Q$$
$$F = \begin{bmatrix} 1 & 0 & 0 \\ 0 & 1 & 0 \\ 0 & 0 & 1 \end{bmatrix} \quad G = \begin{bmatrix} \Delta T & 0 & 0 \\ 0 & \Delta T & 0 \\ 0 & 0 & \Delta T \end{bmatrix} \quad (9)$$

Equation (9) shows the time update through the state and input vector. $\Delta T$ is the unit time. In this study, we set it to 10 Hz according to the LIDAR period.

*2) Measurement Update*

There are two measurements for vehicle localization: 1) the position and heading error correction value determined by NDT scan matching based on lane marking and 2) the result of map matching between the map and extracted points from LIDAR.

$$\eta = [x_t^- + \Delta x_{lane} \ y_t^- + \Delta y_{lane} \ \psi_t^- + \Delta \psi_{lane}]$$
$$r = \sqrt{(x_{LIDAR} - x_t^-)^2 + (y_{LIDAR} - y_t^-)^2} \quad (10)$$
$$\beta = tan^{-1}(\frac{y_{LIDAR} - y_t^-}{x_{LIDAR} - x_t^-}) - \psi_t^-$$

Equation (10) defines the measurements. $\eta$ is obtained by adding the error correction value determined by NDT scan matching to the current state vector. $r$ and $\beta$ are the range and bearing, respectively, between the tunnel facility point extracted from LIDAR and the vehicle position. Thus, the sum of all measurements is as follows:

$$z = [\eta \ r_1 \ \beta_1 \ \cdots \ r_N \ \beta_N]^T \quad (11)$$

$\eta$ is the vehicle pose corrected by matching lane markings. The remaining measurements are the range and bearing of facility points matched by data association. $N$ is the number of extracted tunnel facility points. An observation matrix modeled by linearizing the measurements is required to update the measurements in EKF using $z$.

$$A = \frac{\partial r}{\partial x_t^-} = \frac{-(x_{map} - x_t^-)}{\sqrt{(x_{map} - x_t^-)^2 + (y_{map} - y_t^-)^2}}$$
$$B = \frac{\partial r}{\partial y_t^-} = \frac{-(y_{map} - y_t^-)}{\sqrt{(x_{map} - x_t^-)^2 + (y_{map} - y_t^-)^2}} \quad (12)$$
$$C = \frac{\partial \beta}{\partial x_t^-} = \frac{(y_{map} - y_t^-)}{(x_{map} - x_t^-)^2 + (y_{map} - y_t^-)^2}$$
$$D = \frac{\partial \beta}{\partial y_t^-} = \frac{-(x_{map} - x_t^-)}{(x_{map} - x_t^-)^2 + (y_{map} - y_t^-)^2}$$

Equation (12) is a partial derivative equation used to construct the observation matrix. $A$ and $B$ are derived by partial derivative of the range between the point of the map and the vehicle position into the state vector $\hat{X}_t^-$. Similarly, $C$ and $D$ are the result of the partial derivative of the bearing into the state vector. The observation matrix is constructed based on the partial derivative parameters as follows:

$$H = \frac{\partial h(\hat{X}_t^-, m)}{\partial \hat{X}_t^-} = \begin{bmatrix} 1 & 0 & 0 \\ 0 & 1 & 0 \\ 0 & 0 & 1 \\ A_1 & B_1 & 0 \\ C_1 & D_1 & -1 \\ \vdots & \vdots & \vdots \\ A_N & B_N & 0 \\ C_N & D_N & -1 \end{bmatrix} \quad (13)$$

Equation (13) is the result of constructing the observation matrix $H$. The identity matrix is the Jacobian of $\eta$. The following shows the Jacobian of range and bearing measurements. The vehicle pose is updated when the measurement update is performed by applying the Kalman filter-based on the observation matrix.

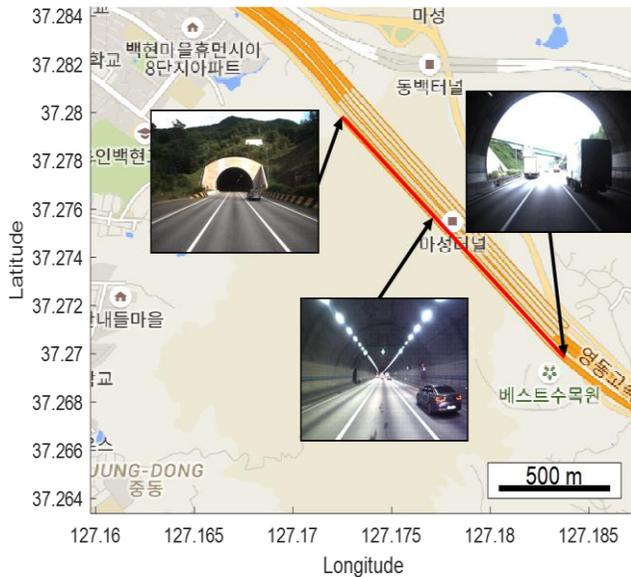

Fig. 20. Experimental area (red line). The experiment was conducted in Masung tunnel, Yeongdong highway in South Korea.

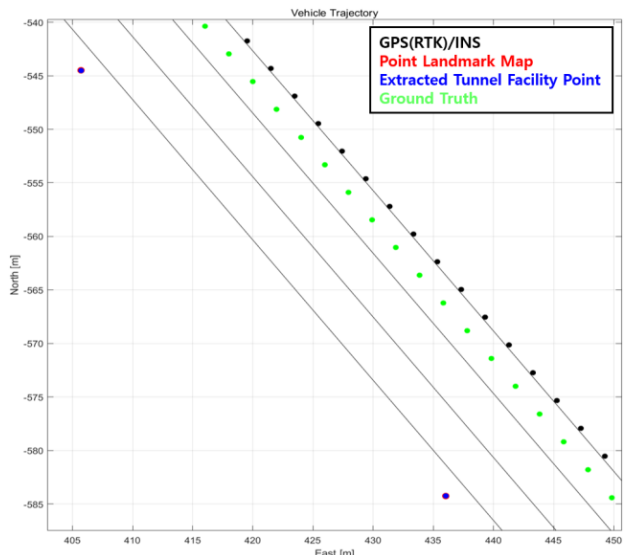

Fig. 21. Ground truth trajectory result obtained using post processed GPS (RTK)/INS trajectory. The drift error of the GPS (RTK)/INS trajectory was corrected by the extracted tunnel facility.

## VI. EXPERIMENTAL RESULT

### A. Experimental Configuration

To verify the tunnel facility-based vehicle localization algorithm, an experiment was conducted at the Masung and Yeongdong highways in South Korea. As shown in Fig. 20, there are three lanes in the tunnel, and the length of the tunnel is 1.5 km. The configuration of the experimental vehicle is shown in Fig. 3. For vehicle localization, we used 3D LIDAR, low cost GPS, and IMU. All algorithms were executed with a mini PC, Intel NUC8i7BEH (CPU: Intel Core i7-8550U @ 2.7 GHz, RAM: 16 GB). In addition, all algorithms were operated in real time based on the robot operating system (ROS).

We conducted the experiment in all three lanes of the tunnel. Then, we verified the tunnel facility point extraction result and the performance of vehicle localization for each lane. In addition, we drove the vehicle at 90–100 km/h for each lane. Through this, we verified whether precise localization was possible under high-speed driving conditions.

We stored all the experimental data in the ROS bag. The ROS bag stores data based on ROS time. When the ROS bag data are played back, data are loaded according to the ROS time flow. Therefore, during playback at 1× speed, localization algorithm operates almost similar to the actual environment. Considering this, we analyzed the real-time operation result by executing the algorithm using the ROS bag data.

We mounted the GPS-RTK and an expensive INS on the experimental vehicle for analyzing the positioning performance. The data obtained from the two sensors were post-processed to determine the tightly coupled GPS (RTK)/INS trajectory. However, because GPS cannot be received in the tunnel, a drift error occurs even when an expensive INS is used. To solve this problem, the drift error was corrected by matching the tunnel facility points extracted using LIDAR and a point landmark map-based on the post-processed GPS (RTK)/INS trajectory.

Fig. 21 shows the ground truth trajectory result. As shown in Fig. 21, the extracted tunnel facility points are matched with the landmark map points. The post-processed GPS (RTK)/INS trajectory is out of the lane because of drift error. Meanwhile, the trajectory of the corrected point landmark map is located at the center of the lane. In this study, we assumed this trajectory to be the ground truth. Then, we analyzed the tunnel facility-based localization performance based on the ground truth.

### B. Analysis of Experimental Result

Fig. 22 shows the result of tunnel facility-based vehicle localization. In the tunnel, the GPS/DR system cannot receive GPS signals from satellites. Therefore, the position is estimated only through DR. As a result, the position error increases owing to the accumulation of drift error. Meanwhile, the proposed method estimates position precisely by correcting the drift error by using tunnel facility-based vehicle localization.

Processing time of tunnel facility based vehicle localization algorithm is shown in Fig. 23. Table VI shows the average and maximum processing time of the tunnel facility-based vehicle localization algorithm. The algorithm is processed rapidly (within 50 ms on an average) for all the lanes, and the maximum processing time is less than 100 ms. Therefore, considering that the LIDAR period is 10 Hz, we conclude that this algorithm can be operated in real time.

In map-aided localization, the extraction of sensor data and map matching requires a considerable amount of time. In tunnel facility-based vehicle localization, tunnel facility points are extracted for map matching. The extraction process requires tunnel wall removal and object clustering. Most of the processing time for tunnel wall removal is consumed for ICP scan matching. However, ICP scan matching is performed with a small number of points by sampling the point clouds around the virtual elliptical cylinder. In addition, the number of points that remain when the tunnel wall is removed is not large. Therefore, object clustering can be processed rapidly, whereby it is possible to extract tunnel facility points rapidly. The map matching is divided into two processes: point landmark matching and NDT scan matching. Point landmark matching is performed rapidly because the number of extracted tunnel facility points is moderate. In the case of NDT scan matching, the number of extracted lane marking points is small. Moreover, the shape of the probability distribution is simple because the lane marking is a solid line. This enables rapid convergence and, in turn, the rapid performance of all the processes, as revealed by the average time of each process in Table VI.

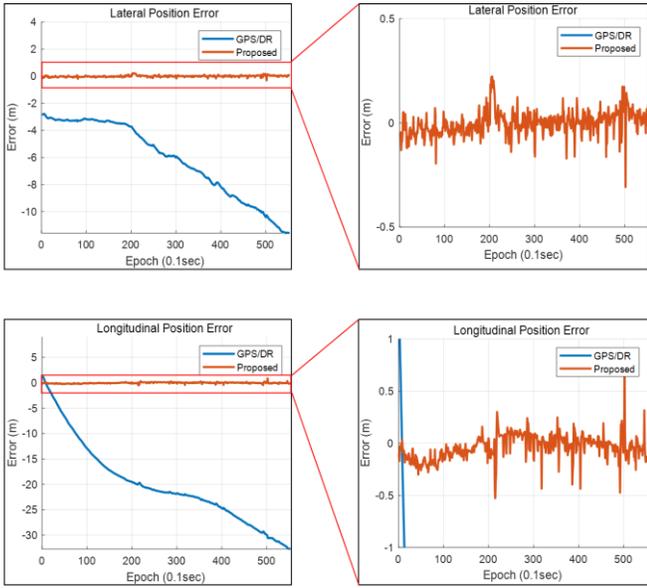

Fig. 22. Comparison of position error between GPS/DR and tunnel facility-based vehicle localization.

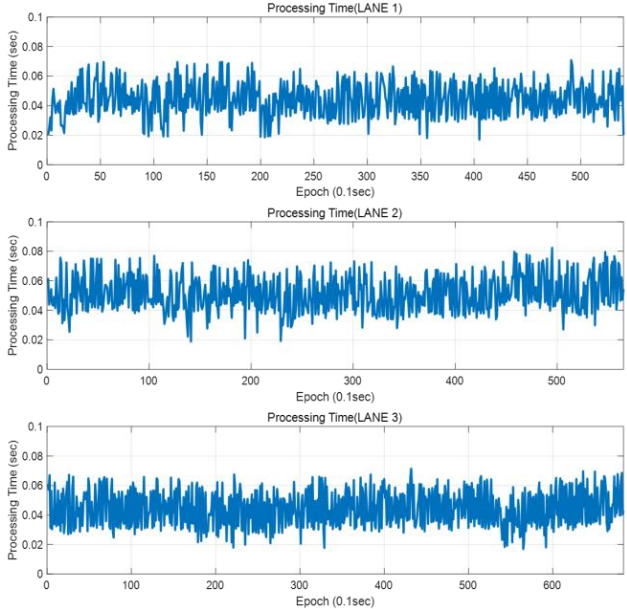

Fig. 23. Processing time of tunnel facility-based vehicle localization algorithm.

TABLE VI
PROCESSING TIME OF TUNNEL FACILITY-BASED
VEHICLE LOCALIZATION ALGORITHM

| Processing Time (ms) | Lane 1 | Lane 2 | Lane 3 |
|---|---|---|---|
| Average time | 44.2 | 51.3 | 44.4 |
| Maximum time | 70.8 | 82.3 | 71.4 |
| **Average Time of Each Process (ms)** | **Lane 1** | **Lane 2** | **Lane 3** |
| Tunnel wall removal time | 13.2 | 13.4 | 14.1 |
| Object clustering | 11.4 | 11.6 | 11.8 |
| Map matching | 18.6 | 25.3 | 17.6 |

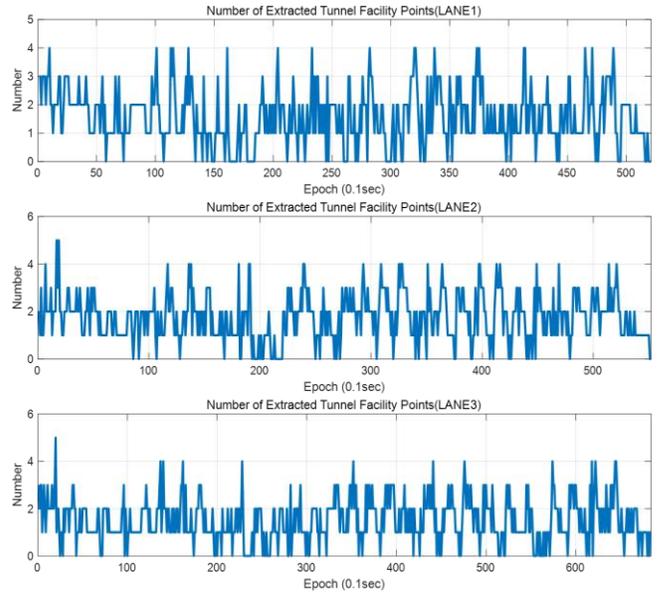

Fig. 24. Result of extracted tunnel facility points for each lane.

TABLE VII
DETECTION RATE OF TUNNEL FACILITY POINTS FOR EACH LANE

| Number of Points (%) | Detection rate of tunnel facility point | | |
|---|---|---|---|
| | Lane 1 | Lane 2 | Lane 3 |
| (N ≥ 1) | 87.3 | 90.9 | 88.9 |
| (N ≥ 2) | 48.6 | 58.3 | 45.9 |
| (N ≥ 3) | 17.5 | 20.5 | 13.5 |
| **Facility Type (%)** | Detection rate of each tunnel facility point | | |
| | Lane 1 | Lane 2 | Lane 3 |
| Fire extinguisher lamp | 83.7 | 84.3 | 87.4 |
| LCS | 1.5 | 2.7 | 1.8 |
| Exit light | 13.9 | 12.6 | 10.7 |
| Exit sign | 0.9 | 0.4 | 0.1 |

Fig. 24 shows the result of the extracted tunnel facility points for each lane. The lane order is from left to right. As shown in Fig. 24, more than one tunnel facility point is extracted in most sections. The detailed tunnel facility point detection rate is shown in Table VII. As shown in the table, most of the tunnel facility points are detected in Lane 2. In addition, the detection rate for three or more tunnel facilities is higher than that for the other lanes. Because Lane 2 is the middle lane in the tunnel and the distance between the fire extinguisher lamp and exit light is similar, it is feasible to extract the two tunnel facilities in a balanced manner. This results in a detection rate that is higher than that of the other lanes. In addition, the detection rate also increases because the distances between the LCS of all the lanes are similar.

With regard to the detection rate by facility type, the fire extinguisher lamp has the highest rate, followed by the exit light. Thus, it is evident that the fire extinguisher lamp is the most important facility in tunnel facility-based vehicle localization, followed by the exit light. A greater number of these two facilities is extracted as compared to the other facilities because of their shorter intervals. However, the detection rate of the exit light is significantly lower than that of the fire extinguisher lamp. There are two reasons for this. First, the probability of the exit light being obscured by other vehicles is high because of its smaller installation height. The second reason is the difference in their shapes. The fire extinguisher is in the form of a triangular column and protrudes from the tunnel wall. Therefore, many points remain even after the removal of tunnel wall points. Meanwhile, the exit light is attached to the tunnel wall. Consequently, it is often removed with the tunnel wall points. As a result, a large difference occurs in the detection rate.

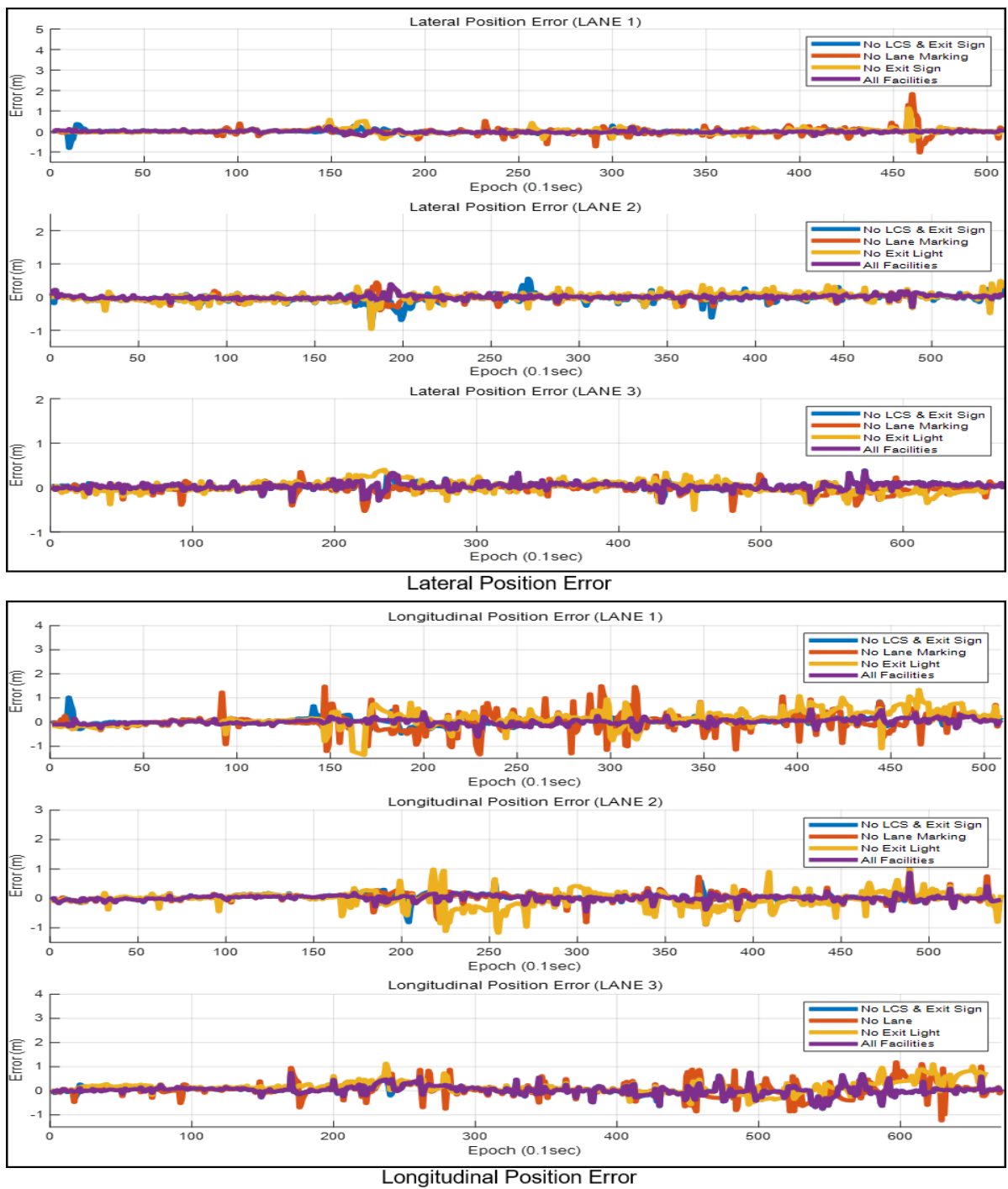

Fig. 25. Position error for each lane according to the usage of tunnel facilities.

TABLE VIII
RMS ERROR FOR EACH LANE ACCORDING TO THE USAGE OF TUNNEL FACILITIES

| RMS Error (m) | Lane 1 | | Lane 2 | | Lane 3 | |
|---|---|---|---|---|---|---|
| | Lateral | Longitudinal | Lateral | Longitudinal | Lateral | Longitudinal |
| No LCS & exit sign | 0.076 | 0.176 | 0.101 | 0.112 | 0.080 | 0.186 |
| No exit light | 0.128 | 0.386 | 0.086 | 0.154 | 0.097 | 0.288 |
| No lane marking | 0.183 | 0.365 | 0.126 | 0.271 | 0.130 | 0.294 |
| All facilities | **0.055** | **0.120** | **0.062** | **0.098** | **0.083** | **0.183** |

The relationship between the tunnel facility detection rate and position estimation performance is illustrated in Table VIII. As shown in the table, the position error is the lowest in Lane 2, which has the highest detection rate. However, it is evident that when we use all the facilities, the position error in all the lanes is estimated to be within 0.1 m and 0.2 m in the lateral and longitudinal directions, respectively. Thus, it is evident that precise vehicle localization is feasible in any lane.

Fig. 25 shows the position error for each lane according to the usage of tunnel facilities. Table VIII shows the RMS error for each lane according to the usage of tunnel facilities. Considering the position error owing to the use of facilities, the best performance is when all the facilities are used. We analyze the vehicle localization performance according to the usage of each tunnel facility. A fire extinguisher lamp is used in all the cases. Vehicle localization is unfeasible if the fire extinguisher lamp is not used, because the number of tunnel facility points extracted is small.

The LCS and exit sign have very low detection rates because of the installation interval. However, the position error is lowered when these facilities are used. These are installed facing the vehicle and are suspended from the tunnel ceiling. Therefore, it is possible for their entire length to be scanned by LIDAR. In addition, there are few false detections. Thus, even if false detection of other facilities occurs, error correction is possible through these facilities.

The exit light is installed on the left wall unlike the fire extinguisher lamp, which is installed on the right wall. In terms of the dilution of precision (DOP), the larger the angle between the detected landmarks, the higher is the position estimation performance [30]. When the exit light is not used, most of the extracted landmark points are located on the right. As a result, DOP becomes very poor. Meanwhile, when we use these two facilities the angle between the landmarks is improved because they are located on opposite walls each other. Consequently, the DOP is improved. In conclusion, the use of two facilities can improve the position estimation performance further.

Because the lane marking in the tunnel is a solid line, it is unfeasible to estimate the longitudinal direction. This is unlike the lane marking comprising dashed lines. However, it enables the estimation of the lateral direction. In addition, it is possible to estimate the heading according to the bending of the tunnel because it is a line parallel to the tunnel. In particular, lane marking is extracted using the LIDAR intensity. Therefore, the risk of false detection is low, and many points can be extracted. Thus, even if an error occurs owing to incorrect detection of other landmark points, it is possible to correct the error to a certain extent.

To summarize, we verify that the drift error of the DR system in the tunnel can be corrected through a tunnel facility-based vehicle localization algorithm. In addition, we verify that the vehicle position is estimated within 0.1 m and 0.2 m in the lateral and longitudinal directions, respectively, in all the lanes. Therefore, tunnel facility-based vehicle localization yields precise results irrespective of where the vehicle is located. Moreover, we verify that all the algorithms can be processed in real time through processing time analysis.

## VII. Conclusion

In this paper, we propose a tunnel facility-based vehicle localization method. First, tunnel facility points are extracted effectively using 3D LIDAR. Next, map-aided localization suitable for highway tunnels is performed based on the tunnel facility point landmark map and lane marking probability distribution map. Thereby, we verify that precise vehicle position estimation in a highway tunnel in real time is possible. The contributions of the proposed algorithm are as follows:

1. A map-aided vehicle localization method suitable for highway tunnels is proposed. Map-aided localization is the most commonly used method for vehicle position estimation. In particular, it is essential in urban canyons or tunnels where it is difficult to receive GPS satellite signals. To date, many types of map-aided vehicle localization methods have been proposed. However, few studies have been conducted on map-aided localization in highway tunnels. The existing map-aided localization using a camera and LIDAR performs map matching based on unique feature points that are contrasted with other surrounding environments. However, highway tunnels mostly comprise elliptical walls. Therefore, it is not possible to extract many feature points for localization in a tunnel and, in turn, reduces position estimation performance in tunnels when we use existing map-aided vehicle localization method. To solve this problem, we performed vehicle localization based on point landmark. Unlike existing map-aided localization, point landmark-based localization does not require many points. This is because point landmark map matching matches the representative points of an object. As shown in [13], position estimation is feasible even if there is only one landmark point. In addition, the tunnel facility used for the point landmark has a set standard and installation location. This facilitates the extraction of point landmarks through a tunnel facility. Therefore, this method is suitable for vehicle localization in highway tunnels where the feature points that can be extracted are few in number.

2. Vehicle localization is feasible without the installation of additional infrastructure in the tunnel. Most of the studies on vehicle localization in highway tunnels estimated vehicle position through V2V or V2I using UWB, which requires the installation of additional infrastructure (such as UWB for localization). Meanwhile, it is mandatory to install (during tunnel construction and according to the related manual) the tunnel facilities used in this study. Therefore, vehicle localization is feasible without the need for installing additional infrastructure. In addition, because there are very few tunnels installed with UWB, most of the UWB-based vehicle localization is dependent on the simulation results. Meanwhile, tunnel facility-based vehicle localization can be verified conveniently through actual experiments if only maps are available. Therefore, it is convenient and inexpensive to apply it to real autonomous vehicles.

3. Precise vehicle localization is possible with very small-sized maps such as point landmarks and probability distribution maps. The point landmark map stores only the positions of the representative points of an object. In addition, probability distribution maps are smaller in size than other maps because only the mean and covariance of probability distributions are stored in these. The map size for vehicle localization is based on the 1.5 km tunnel, the size of the point landmark map is 6.4 kb, and that of the lane marking probability distribution map is 53.4 kb. The sum of these is 59.8 kb. As a result, map-aided localization is feasible with a very small map size. In addition, the experimental results verify that highly precise vehicle localization is feasible within 0.1 m and 0.2 m in the lateral and longitudinal directions, respectively.

4. It was verified that real-time position estimation is feasible at high speeds. Real-time estimation is essential for autonomous vehicle navigation, particularly in the case of highways because the vehicle speed is excessively fast. This study verified that our localization algorithm operates highly rapidly. 3D LIDAR scans a 3D environment and provides a precise point cloud. Therefore, LIDAR-based localization consumes a substantial amount of time to extract feature points from a point cloud or map matching. Meanwhile, our proposed method can decrease the process time. The tunnel is composed mostly of a wall. Therefore, only a few points remain when the wall is removed. This shortens the time required for tunnel facility point extraction and map matching. Thus, the processing speed is high, and real-time processing is feasible.

Using the proposed vehicle localization algorithm, we verified that precise vehicle position estimation in the tunnel in real time is possible. Our research is focused on vehicle localization in highway tunnels. Therefore, the initial position is acquired from the GPS/DR. In the future, we aim to extend this work beyond tunnels. We plan to design an algorithm that enables map-aided vehicle localization in an entire highway section and verify its performance.

ACKNOWLEDGMENT

This research was supported by a grant (19TLRP-B101406-05) from the Transportation & Logistics Research Program funded by Ministry of Land, Infrastructure and Transport of the Korean government.

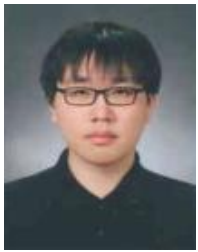
**Kyuwon Kim** received the B.S. and M.S. degrees from Department of Electrical and Electronics Engineering at Konkuk University, Seoul, South Korea, in 2015 and 2017, respectively. He has been a Ph.D. student of Navigation and Control System Laboratory at Konkuk University since 2017. His research interests are autonomous vehicle navigation, UAV navigation, sensor fusion SLAM, map-aided localization, and indoor navigation.

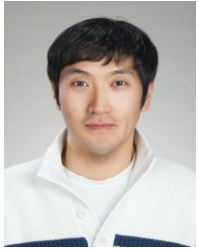
**Junhyuck Im** received the B.S. degree from Department of Electronic Engineering at Hongik University, Seoul, South Korea, in 2008. He received the M.S and Ph.D. degrees from Department of Electrical and Electronics Engineering at Konkuk University, Seoul, South Korea, in 2011 and 2017, respectively. He has been a senior researcher at Korea Land and Geospatial Informatix Corporation since 2018. His research interests are LIDAR-based localization, self-driving car, SLAM, indoor positioning, and Loran-C signal processing.

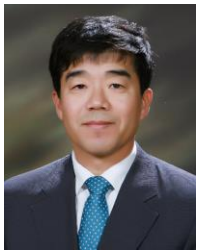
**Gyuin Jee** (M') received the B.S. and M.S. degrees from Department of Control and Instrumentation Engineering at Seoul National University, Seoul, South Korea, in 1982 and 1984, respectively. He received the Ph.D. degree from Department of System and Control Engineering at Case Western Reserve University, USA, in 1989. He has been a professor at Konkuk University since 1992. His research interests are GNSS receiver signal processing, GPS jamming, anti GPS spoofing, SLAM, unmanned vehicle navigation, and multiple sensors-based navigation.